\documentclass[runningheads]{llncs}
\usepackage{graphicx}
\usepackage{comment}
\usepackage{amsmath,amssymb} 
\usepackage{color}
 \usepackage{multirow}

\newcommand*\samethanks[1][\value{footnote}]{\footnotemark[#1]}

\usepackage{float}
\newfloat{figtab}{htb}{fgtb}
\makeatletter
  \newcommand\figcaption{\def\@captype{figure}\caption}
  \newcommand\tabcaption{\def\@captype{table}\caption}
\makeatother

\begin{document}
\pagestyle{headings}
\mainmatter
\def\ECCVSubNumber{2640}  

\title{GINet: Graph Interaction Network for Scene Parsing} 

\titlerunning{Graph Interaction Network for Scene Parsing} 
\authorrunning{Tianyi Wu et al.} 
\author{Tianyi Wu\inst{1,2}\thanks{Equal contribution $^{*}$\ Corresponding author} \and Yu Lu\inst{3}\samethanks \and Yu Zhu\inst{1,2}  \and \\
 Chuang Zhang\inst{3} \and MingWu\inst{3} \and Zhanyu Ma\inst{3} \and Guodong Guo\inst{1,2}$^{*}$}
\institute{{Institute of Deep Learning, Baidu Research, Beijing, China \\ \email{\{wutianyi01, zhuyu05, guoguodong01\}@baidu.com}}\\
\and
{National Engineering Laboratory for Deep Learning Technology and Application, Beijing, China} \\
\and
{Beijing University of Posts and Telecommunications, Beijing, China \\ \email{\{aniki, zhangchuang, wuming, mazhanyu\}@bupt.edu.cn}}} 

\maketitle

\begin{abstract}
   Recently, context reasoning using image regions beyond local convolution has shown great potential for scene parsing. In this work, we explore how to incorperate the linguistic knowledge to promote context reasoning over image regions by proposing a Graph Interaction unit (GI unit) and a Semantic Context Loss (SC-loss). 
   The GI unit is capable of enhancing feature representations of convolution networks over high-level semantics and learning the semantic coherency adaptively to each sample. 
   Specifically, the dataset-based linguistic knowledge is first incorporated in the GI unit to promote context reasoning over the visual graph, then the evolved representations of the visual graph are mapped to each local representation to enhance the discriminated capability for scene parsing. 
   GI unit is further improved by the SC-loss to enhance the semantic representations over the exemplar-based semantic graph. 
   We perform full ablation studies to demonstrate the effectiveness of each component in our approach. 
   Particularly, the proposed GINet outperforms the state-of-the-art approaches on the popular benchmarks, including Pascal-Context and COCO Stuff. 
\end{abstract}

\section{Introduction}

Scene parsing is a fundamental and challenging task with great potential values in various applications, such as robotic sensing and image editing. 
It aims at classifying each pixel in an image to a specified semantic category, including objects (\emph{e.g.}, bicycle, car, people) and stuff (\emph{e.g.}, road, bench, sky). 
Modeling context information is essential for scene understanding \cite{belongie2002shape,tu2005image,biederman1982scene}.
Since the work by Long \cite{Long_2015_CVPR} with fully convolutional networks (FCN), it has attracted more and more attention for context modeling in semantic segmentation or scene parsing.

Early works are some approaches that lie in the stack of local convolutions to capture the context information. Several works employed dilation convolution \cite{chen2014semantic,yu2015multi,chen2017deeplab,chen2017rethinking,chen2018encoder,wu2019consensus,wu2019tree}, kronecker convolution \cite{wu2019tree} and pooling operations \cite{liu2015parsenet,zhao2017pyramid} to obtain a wider context information. Recent works \cite{yuan2018ocnet,fu2019dual,zhang2019co} introduced non-local operations \cite{Wang2017Non} to integrate the local feature with their contextual dependencies adaptively to capture richer contextual information. 
Later, several approaches \cite{huang2018ccnet,li2019expectation,zhu2019asymmetric} were proposed to reduce the computation of non-local operations.
More recently, using image regions for context reasoning
\cite{chen2019graph,li2018beyond,liang2018symbolic,zhang2019latentgnn} has shown great potential for scene parsing. 
These methods were proposed to learn a graph representation from visual features, where the vertices in the graph define clusters of pixels (``region''), and edges indicate the similarity or relation between these regions in the feature space. In this way, contextual reasoning can be performed in the interaction graph space, then the evolved graph is projected back to the original space to enhance the local representations for scene parsing.

\begin{figure}[t]
   \begin{center}
      \includegraphics[scale = 0.3]{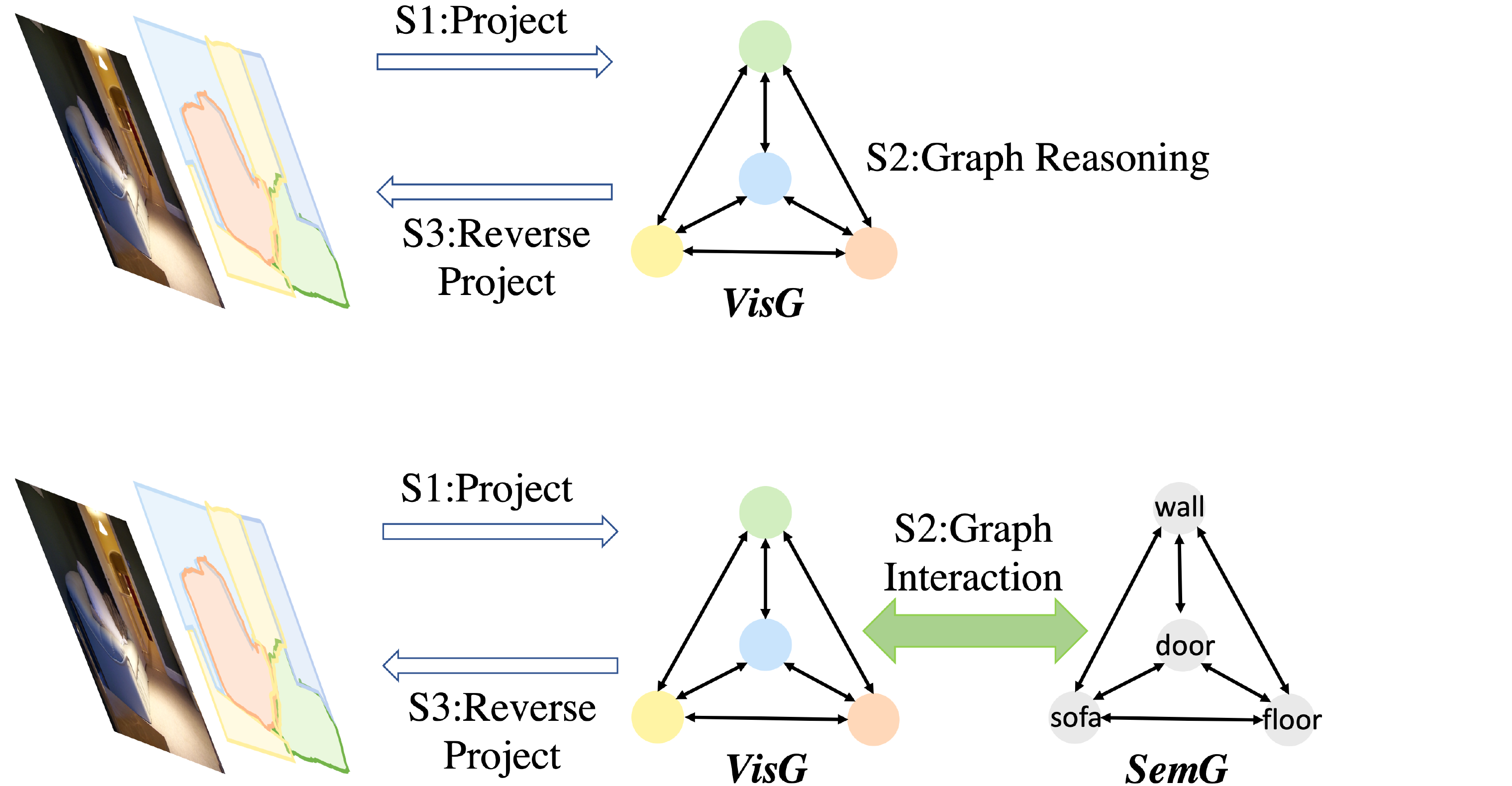}
   \end{center}
      \caption{Comparison of different contextual reasoning frameworks. {\em Top}: Contextual reasoning over the visual graph.
       {\em Bottom}: Our proposed Graph Interaction Network(GINet). 
       Note: {\em VisG}: visual graph. {\em SemG}: semantic graph.
       }
   \label{fig:example}
   \label{fig:onecol}
 \end{figure}
 
In this paper, instead of solely performing context reasoning over the visual graph representation for 2D input images or visual features (as shown in the top of Figure~\ref{fig:example}), we seek to incorporate linguistic knowledge, such as linguistic correlation and label dependency, to share the external semantic information across locations that can promote context reasoning over the visual graph. Specifically, we propose a Graph Interaction unit (GI unit), which first incorporates the dataset-based linguistic knowledge into feature representation over the visual graph, and re-projects the evolved representations of the visual graph back into each location representation for enhancing the discriminative capability (as shown in the bottom of Figure~\ref{fig:example}). Intuitively, the external knowledge is modeled as a semantic graph which is formed as vertices with linguistic entities (e.g., cup, table and desk) and edges with entity relationships (e.g., semantic hierarchy, concurrence and spatial interactions). GI unit shows the interaction between the visual and semantic graph. Furthermore, we introduce a Semantic Context Loss, which aims at learning an exemplar-based semantic graph to better represents the sample adaptively, where the categories that appear in the scene are emphasized while those do not appear in the scene are suppressed. 
Details of the proposed method are presented in Section~\ref{section:3}.

The most relevant works to our approach are \cite{chen2019graph,liang2018symbolic,li2018beyond}.
Liang \cite{liang2018symbolic} proposed a Symbolic Graph Reasoning layer to perform reasoning over a group of symbolic nodes. The SGR explores how to harness various external human knowledge for endowing the networks with the capability of semantic global reasoning. In contrast, our method explores how to incorporate a dataset-based linguistic knowledge to promote context reasoning over image regions.
Li \cite{li2018beyond} proposed a Graph Convolutional Unit to project a 2D feature map into a sample-dependent graph structure by assigning pixels to the vertices of the graph and learning a primitive grouping of scene components. 
Chen \cite{chen2019graph} introduced the Global Reasoning unit for reasoning globally, which projects information from the coordinate space to nodes in an interactive space graph to directly reason over globally-aware discriminative features.
Different from these approaches, we propose to reason over the visual graph and the prior semantic graph. 
The semantic graph is employed to promote contextual reasoning and lead the generation of the exemplar-based semantic graph from the visual graph.

We conduct extensive experiments on different challenging datasets to validate the advantages of the proposed GI unit and SC-loss for scene parsing. Meanwhile, ablation studies are performed to demonstrate the effectiveness of each component in our approach. Experimental results are shown in Section~\ref{section:4}. 

The main contributions of this work include:
 
\begin{itemize}
\item[$\bullet$] A novel Graph Interaction unit (GI unit) is proposed for contextual modeling, which incorporates the dataset-based linguistic knowledge for promoting context reasoning over the visual graph. Moreover, it learns an exemplar-based semantic graph as well.
\item[$\bullet$] A Semantic Context Loss (SC-loss) is proposed to regularize the training procedure in our approach, which emphasizes the categories that appear in the scene and suppresses those do not appear in the scene.
\item[$\bullet$] A Graph Interaction Network (GINet) is developed, based on the proposed GI unit and SC-loss for scene parsing; It provides significant gains in performance over the state-of-the-art approaches on Pascal-Context \cite{Mottaghi_2014_CVPR} and COCO Stuff \cite{caesar2018coco}, and achieves a competitive performance on ADE20K dataset \cite{Zhou_2017_CVPR}. 
\end{itemize}
   
\section{Related work}

In this section, we briefly overview the recent progress in contextual modeling for scene parsing. They can be mainly divided into two categories based on whether graph reasoning is considered.

There are several model variants of FCN \cite{Long_2015_CVPR} proposed to exploit the contextual information.
Some methods \cite{eigen2015predicting,lin2016efficient,wang2018multiscale,chen2017deeplab,chen2017rethinking,yuan2018ocnet,Wang_2020_CVPR,zhao2017pyramid,wang2019ls,yuan2019object} 
were proposed to learn the multi-scale contextual information.
DeepLabv2 \cite{chen2017deeplab} and DeepLabv3 \cite{chen2017rethinking} utilized an atrous spatial pyramid pooling to capture contextual information, which consists of parallel dilation convolutions with different dilation rates. 
TKCN \cite{wu2019tree} introduced a tree-structured feature aggregation module for encoding hierarchical contextual information.
The pyramid pooling module is proposed by PSPNet \cite{zhao2017pyramid} to collect the effective contextual prior, containing information on different scales.
Moreover, the encoder-decoder structures \cite{zhou2018d,chaurasia2017linknet,badrinarayanan2017segnet,pohlen2017full,peng2017large} based on UNet \cite{ronneberger2015u} fuse the high-level and mid-level features to obtain context information.
DeepLabV3+ \cite{chen2018encoder} combines the properties of the above two methods that add a decoder upon DeepLabV3 to help model obtain multi-level contextual information and preserve spatial information.
Differently, CGNet \cite{wu2018cgnet} proposed a Context Guided block for learning the joint representation of both local features and surrounding context.
In addition, inspired by ParseNet \cite{liu2015parsenet}, a global scene context was utilized in some methods \cite{yu2018learning,zhao2018psanet} by introducing a global context branch in the network.
EncNet \cite{zhang2018context} introduced Encoding Module to capture the global semantic context and predict scaling factors to selectively highlight feature maps.
Recently, there were many efforts \cite{fu2019dual,yuan2018ocnet,zhang2019co} to break the local limitations of the convolution operators by introducing the Non-local block \cite{Wang2017Non} into the feature representation learning to capture spatial context information.
Furthermore, some methods \cite{li2019expectation,zhu2019asymmetric,huang2018ccnet} proposed to reduce the computational complexity of Non-Local operations. 
More recently, SPGNet \cite{cheng2019spgnet} proposed a Semantic Prediction Guidance module which learns to re-weight the local features through the guidance from pixel-wise semantic prediction.

Some other methods introduced a graph propagation mechanism into the CNN network to capture a more extensive range of information. 
GCU \cite{li2018beyond} got inspiration from region-based recognition and presented a graph-based representation on semantic segmentation and object detection tasks.
GloRe \cite{chen2019graph} and LatenGNN \cite{zhang2019latentgnn} performed a global relation reasoning  by aggregating features with similar semantics to an interactive space.
SGR \cite{liang2018symbolic} extracted the representation nodes of each category from the features and use external knowledge structures to reason about the relationship between categories.
These methods have typical projection, reasoning, and back projection steps. Based on these steps, our approach further promotes graph reasoning by incorporating semantic knowledge.
Finally, Graphonomy and GraphML \cite{gong2019graphonomy,DBLP:journals/corr/abs-1911-12053} proposed to propogate graph features of different datasets to unify the human parsing task.
However, our approach explores the correlation between visual and semantic graph to facilitates the context modeling capabilities of the model.

\section{Approach}
\label{section:3}
In this section, we first introduce the framework of the proposed Graph Interaction Network (GINet). 
Then we present the design of the Graph Interaction unit (GI unit) in details. Finally, we give a detailed description of the proposed Semantic Context loss (SC-loss) for the GINet.

\subsection{Framework of Graph Interaction Network (GINet)}
\begin{figure*}
   \begin{center}
   \includegraphics[scale = 0.45]{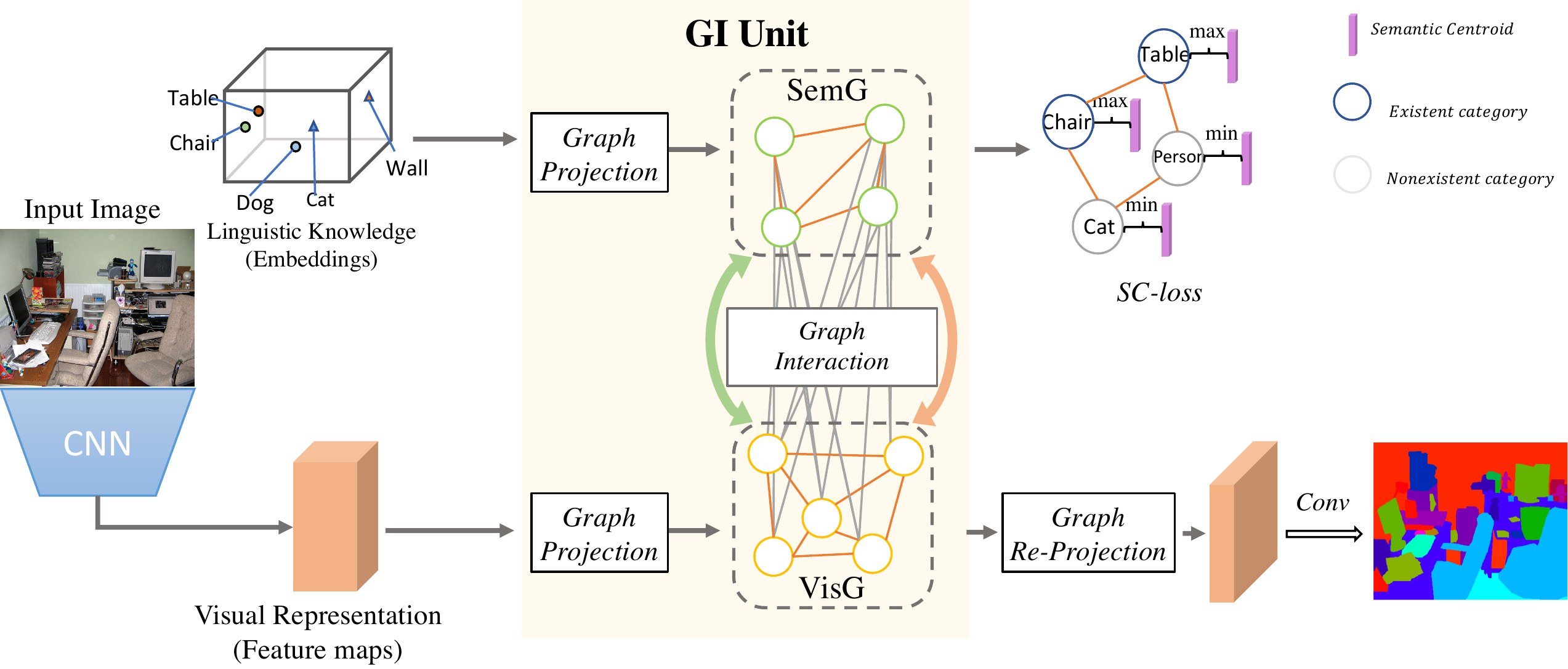}
   \end{center}
     \caption{The overall framework of the proposed Graph Interaction Network (GINet). 
    ({\em Best viewed in color.})}
   \label{fig:framework}
   \end{figure*}   

Different from previous methods that only perform contextual reasoning over the visual graph built on visual features \cite{li2018beyond,chen2019graph}, our GINet facilitates the graph reasoning by incorporating semantic knowledge to enhance the visual representations. 
The proposed framework is illustrated in Figure~\ref{fig:framework}. Firstly, we adopted a pre-trained ResNet \cite{he2016deep} as the backbone network, where visual features can be extracted given an input 2D image. Meanwhile, the dataset-based linguistic knowledge can be extracted in the form of categorical entities (classes), which is fed into word embedding (e.g., GloVe \cite{pennington2014glove}) to achieve semantic representations. Secondly, visual features and semantic embedding representations are passed by the graph projection operations in the proposed GI unit to construct two graphs, respectively. A detailed definitions of graph projection operations are presented in Section \ref{Graph_Construction}. Accordingly, one graph that encodes dependencies between visual areas is built over visual features, where nodes indicate visual regions and edges represent the similarity or relation between those regions. The other graph is built over the dataset-dependent categories (represented by word embeddings), which encodes the linguistic correlation and label dependency. Next, a graph interaction operation is processed in the GI unit, where the semantic graph is employed to promote contextual reasoning over the visual graph and guide the generation of the exemplar-based semantic graph extracted from the visual graph. 
Then, the evolved visual graph generated by the GI unit is passed by Graph Re-projection operation for enhancing the discriminative ability for each local visual representation, while the semantic graph is updated and constrained by the Semantic Context loss during the training phase. Finally, we employ an $1\times1$ Conv followed by a simple bilinear upsampling to obtain the parsing results.
   
\subsection{Graph Interaction Unit}

The goal of the proposed GI unit is to incorporate dataset-based linguistic knowledge for promoting the local representations. First, the GI unit takes visual and semantic representations as inputs, conduct contextual reasoning by generating a visual graph and a semantic graph. 
Second, a graph interaction is performed between the two graphs to evolve node features by the guidance of similarity of visual nodes and semantic nodes. 

{\bf Graph Construction:} \

The first step is to define the projection that maps the original visual and semantic features to an interaction space. Formally, given the visual feature maps $X\in \mathbb{R}^{L \times C}$, where $L=H \times W$, $H$ and $W$ indicate the height and width of the feature map, and $C$ is the channel dimensions. We aim to construct a visual graph representation $P \in \mathbb{R}^{N \times D}$, where $N$ is the number of nodes in the visual graph and $D$ is the desired feature dimension for each node. Inspired by the works \cite{li2019graph,gong2019graphonomy}, we introduce a transformation matrix $Z \in \mathbb{R}^{N \times L}$ that projects the local presentation $X$ to a high-level graph representation $P$, which can be computed as follows:
\begin{equation}
P = ZXW,
\end{equation}
where $W \in \mathbb{R}^{C \times D}$ is introduced as trainable parameters to convert the feature dimension from $C$ to $D$, and $Z$ adaptively aggregates local features to a node in the visual graph. 

Next, we define a dataset-dependent semantic graph. Particularly, we aim to build a semantic graph presentation $S \in \mathbb{R}^{M \times D}$ over the object categories of a specific dataset, to encode the linguistic correlation and label dependency. $M$ denotes the number of nodes, which is equal to the number of categorical entities (classes) in the dataset. $D$ is the feature dimension for each node in the semantic graph. 
Specifically, we first use the off-the-shelf word vectors \cite{pennington2014glove} to get semantic representation $l_{i} \in \mathbb{R}^{K}$ for each category $i \in \{0,1,..,M-1\}$, $K=300$.
Then, a MLP layer is employed to adjust the linguistic embedding to suit the reasoning with the visual graph. This transformation process can be formulated as follow:
\begin{equation}
S_{i} = MLP(l_{i}),  i \in \{0,1,..,M-1\},
\end{equation}
where $S_{i}$ represents node features for each category.
  
{\bf Graph Interaction:}\label{Graph_Construction}
Next, we present how our GI unit incorporates dataset-based linguistic knowledge for promoting context reasoning and extracting the exemplar-based semantic graph from the visual graph.
For simplicity, we abbreviate the visual graph and semantic graph as VisG and SemG, respectively.
We first evolve both graphs separately and then perform the interaction between graphs.
Then SemG and VisG attentively propagate information interatively, including 1) Semantic to Visual (S2V), and 2) Visual to Semantic (V2S).

Specifically, we first perform graph convolution \cite{Kipf2016Semi} on the VisG to get evolved graph representation $ \widetilde{P}$  that is suitable for interacting with the SemG. This process can be formulated as follows:
\begin{equation}
\widetilde{P} =  f((A_{v}+I) P W_{v}),
\end{equation}
where the adjacency matrix $A_{v} \in \mathbb{R}^{N \times N}$ is randomly initialized and updated by the gradient descent; $I $ is an identity matrix;  $W_{v} \in \mathbb{R}^{D \times\ D}$ are trainable parameters; and $f$ is a nonlinear activation function.
Through reasoning over the VisG, we can update node representation to capture more visual context information and to interact with SemG. 
Next we perform similar graph convolution \cite{Kipf2016Semi} on the SemG for adjusting the representation of SemG, according to:
\begin{equation}
\widetilde{S} = f ((A_{s}+I) S W_{s}),
\end{equation}
where $\widetilde{S}$ indicates an updated the graph representation of SemG; $A_{s} \in \mathbb{R}^{M \times M}$ is a learnable adjacency matrix or co-occurrence matrix that represents connections between semantic correlation or label dependency; 
$W_{s} \in \mathbb{R}^{D \times\ D}$ are trainable parameters.
By performing a propagation of feature information from neighboring nodes, we can improve the representation of each semantic node.
   
In S2V step: we utilize the evolved SemG to promote contextual reasoning over the VisG $\widetilde{P}$.
Specifically, to explore the relationship between two nodes from VisG and SemG, we compute their feature similarity as a guidance matrix $G^{s2v} \in \mathbb{R}^{N \times M}$.
For one node $\widetilde{p_i} \in \mathbb{R}^{ D}$ in the  VisG $\widetilde{P} $ and one node $\widetilde{s_j} \in \mathbb{R}^{D}$ in the SemG,
we can compute the guide information $G^{s2v}_{i, j}$ that represents the assignment weight of the node $\widetilde{s_j}$ in SemG  to the node $ \widetilde{p_i} $ in VisG as follows:
\begin{equation}
G_{i, j}^{s2v}=\frac{\exp \left(W_{p}\widetilde{p_{i}} \cdot W_{s}\widetilde{s_{j}}\right)}{\sum\nolimits_{m=1}^{M} \exp \left(W_{p}\widetilde{p_{i}} \cdot W_{s}\widetilde{s_{m}}\right)}\label{G_s2v},
\end{equation}
where $i \in \{1,...,N\}, j \in \{1,...,M\} $ and $W_{p} \in \mathbb{R}^{D/2 \times D}$ and $W_{s}\in \mathbb{R}^{D/2 \times D}$ are a learnable matrix to further reduce the feature dimension.
After obtaining the guidance matrix $G^{s2v}$, we can distill information from SemG to enhance the representation of VisG, according to :
\begin{equation}
P_{o} = \widetilde{P} + \beta_{s2v} G^{s2v}\widetilde{S}W_{s2v}\label{s2v},
\end{equation}
where $W_{s2v} \in \mathbb{R}^{D \times D}$ is a trainable weight matrix, $\beta_{s2v} \in \mathbb{R}^{N}$ is a learnable vector with zero initialization and can be updated by a standard gradient decent.
We use a simple sum to melt information from graphs, which may be alternatively replaced by other commutative operators such as mean, max, or concatenate.
With the help of the guidance matrix $G_{s2v}$, we effectively constructed the correlation between visual regions and semantic concepts, and incorporate corresponding semantic features into the visual node representation.

In V2S step: we adopt a similar method elaborated in Equation(\ref{G_s2v}) to obtain the guidance matrix $G_{v2s} \in \mathbb{R}^{M \times N}$.
Formally, the guide information $G^{v2s}_{i, j}$ that can be calculated as follows:
\begin{equation}
G_{i, j}^{v2s}=\frac{\exp \left(W_{s}\widetilde{s_{i}} \cdot W_{p}\widetilde{p_{j}}\right)}{\sum\nolimits_{n=1}^{N} \exp \left(W_{s}\widetilde{s_{i}} \cdot W_{p}\widetilde{p_{n}}\right)},
\end{equation}
where $i \in \{1,...,M\}, j \in \{1,...,N\} $.
After getting the guidance matrix $G^{v2s}$, we update the graph representation of the SemG for generating the exemplar-based SemG according to:
\begin{equation}
S_{o} = \beta_{v2s} \widetilde{S} + G^{v2s}\widetilde{P}W_{v2s}\label{v2s},
\end{equation}
where $W_{v2s} \in \mathbb{R}^{D \times D}$ is a trainable weight matrix, $\beta_{v2s} \in \mathbb{R}^{M}$ is a learnable vector and initialized by zeros.
We extract the exemplar-based semantic graph from VisG with the guidance matrix $G_{v2s}$.
By combining the S2V and V2S steps, the proposed GI unit enables the whole model to learn more discriminative features for performing fine pixel-wise classification and generate a semantic graph for each input image.

{\bf Unit outputs:} \ 
The GI unit has two outputs, one is the exemplar-based SemG, which are described in detail in section \ref{SC_loss}, and the other is the VisG enhanced by semantic information.
The evolved node representation of VisG can be used to enhance the discriminative ability of each pixel feature further.
As previous methods \cite{li2019graph,li2018beyond}, we reuse projection matrix $Z$ to reverse project the $P_{o}$ to 2D pixel features.
Formally, Given node features $P_{o} \in \mathbb{R}^{N \times D}$ of the VisG, the reverse projection (or Graph Re-Projection) can be formulated as follows:
\begin{equation}
\widetilde{X} = Z^{T}P_{o}W_{o} + X,
\end{equation}
where $W_{o} \in \mathbb{R}^{D \times C} $ is a trainable weight matrix that transform the node representation from $\mathbb{R}^{D}$ to $\mathbb{R}^{C}$, $Z^{T} \in \mathbb{R}^{L \times N}$ means the transposed matrix of $Z$,
and we employ a residual connection \cite{he2016deep} to promote the gradient propagation during training.

\subsection{Semantic Context Loss}\label{SC_loss}
\label{subsection:34}
We propose a Semantic Context Loss or simply SC-loss to constrain the generation of exemplar-based SemG. 
It emphasizes the categories that appear in the scene and suppresses those do not appear in the scene, which makes the GINet a capable of enhancing the external semantic knowledge adaptively to each sample.
Specifically, we first define a learnable semantic centroid $c_{i} \in \mathbb{R}^{D}$ for each category. Then, for each semantic node $\widetilde{s_{i}} \in \mathbb{R}^{D}$ in a SemG $\widetilde{S_o}$, 
we compute a score $v_{i}$ by performing a simple dot product with a sigmoid activation upon $\widetilde{s_{i}}$ and $c_{i}$. The $v_{i}$ ranges from 0 to 1 and is trained with the BCE loss.
The SC-loss minimizes the similarity between the node feature in the semantic graph and the semantic centroid of nonexistent categories, and maximizes the similarity to existent classes.
If $v_i$ is closer to 1, the corresponding category exists in the current sample; Otherwise, it does not exist. 
The SC-loss can be formulated as follows:
\begin{equation}
Loss_{sc} = - \frac{1}{M} \sum\nolimits_{i=1}^{M} (y_{i} \cdot logv_{i})+(1-y_{i}) log(1-v_{i}),
\end{equation}
where $y_{i} \in \{0,1\}$ represents the presence of each category in ground truth. 
The proposed SC-loss is different from SE-loss in EncNet \cite{zhang2018context}. It built an additional
fully connected layer on top of the Encoding Layer \cite{zhang2018context} to make individual predictions and was employed to improve the parsing of small objects. However, our SC-loss is employed to improve the generation of exemplar-based SemG.

We also add a full convolution segmentation head attached to Res4 of the backbone to obtain the segmentation result.
Therefore, the objective of the GINet consists of a SC-loss, an auxiliary loss, and a cross-entropy loss, which can be formulated as:
\begin{equation}
Loss = \lambda Loss_{sc} + \alpha Loss_{aux} + Loss_{ce},
\end{equation}
where $\lambda$ and $\alpha$ are hyper-parameters, the selection of $\lambda$ is discussed in the experiment section, and $\alpha$ for auxiliary loss is set to 0.4 similar some previous methods \cite{zhang2018context,zhang2019co,fu2019dual,wu2019fastfcn}.

\section{Experiments}
\label{section:4}

In this section, we perform a series of experiment to evaluate the effectiveness of our proposed Graph Interaction unit and SC-loss. 
Firstly, we give an introduction of the datasets that are used for scene parsing, i.e., Pascal-Context \cite{Mottaghi_2014_CVPR}, COCO Stuff \cite{caesar2018coco}, and ADE20K \cite{Zhou_2017_CVPR}.
Next, we conduct extensive evaluations with ablation studies of our proposed method on these datasets.

\subsection{Datasets} 
{\bf Pascal-Context} \cite{Mottaghi_2014_CVPR} is a classic set of annotations for PASCAL VOC2010, which has 10,103 images. 
In the training set, there are 4,998 images. The remaining 5,105 images form the validation set.
Following previous works \cite{zhang2018context,zhang2019co,Zhou_2017_CVPR}, 
we use the same 59 most frequent categories along with one background category(60 in total) in our experiments.

{\bf COCO Stuff} \cite{caesar2018coco} has a total number of 10,000 images with 183 classes including an 'unlabeled' class,
where 9,000 images are used for training while 1,000 images for validation. We follow the same settings as in \cite{fu2019dual,li2019expectation},
the results are reported on the data contains 171 categories (80 objects and 91 stuff) annotated for each pixel.

{\bf ADE20K} \cite{Zhou_2017_CVPR} is a large scale dataset for scene parsing with 25,000 images and 151 categories.
The dataset is split into the training set, validation set and test set with 20,000, 2,000, and 3,000 images, respectively.
Following the standard benchmark \cite{Zhou_2017_CVPR}, we validate our method on 150 categories, where the background class is not included.
\subsection{Implementation Details} 
During training, we use ResNet-101 \cite{he2016deep} (pre-trained on ImageNet) as our backbone.
For retaining the resolution of the feature map, we use the Joint Pyramid Upsampling Module \cite{wu2019fastfcn} instead of the dilation convolution for saving the training time, resulting in stride-8 models. 
We empirically set the number of nodes in VisG as 64, and node dimensions as 256.
Similar to prior works \cite{chen2017rethinking,zhang2018context}, we employ a poly learning rate 
policy \cite{chen2017deeplab} where the initial learning rate is updated by $lr=base\_lr*(1-\frac{iter}{total\_iter})^{0.9}$ after each iteration.
The SGD \cite{sutskever2013importance} optimizer is applied with 0.9 momentum and 1e-4 weight decay.
The input size for all datasets is set to 520 $\times$ 520.
 For data augmentation, we apply random flip, random crop, and random scale (0.5 to 2) using the zero-padding if needed. 
The batch size is set to 16 for all datasets. 
We set the initial learning rate to 0.005 on ADE20K dataset and 0.001 for others.
The networks are trained for 30k, 150k, 100k iterations on Pascal-Context \cite{Mottaghi_2014_CVPR}, ADE20K \cite{Zhou_2017_CVPR}, COCO Stuff dataset \cite{caesar2018coco}, respectively.

During the validation phase, we follow \cite{zhang2018context,zhang2019co,wu2019fastfcn} to average the multi-scale \{0.5, 0.75, 1.0, 1.25, 1.5, 1.75\} predictions of network.
The performance is measured by the standard mean intersection of union (mIoU) in all experiments.


\begin{figure}[t]
   \begin{minipage}{0.5\linewidth}
    \centering
       \makeatletter\def\@captype{table}\makeatother\caption{Ablation study on PASCAL-Context dataset. ``GI" indicates Graph Interaction Unit. ``SC-loss" represents Semantic Context Loss. 
        ``VisG" means that context reasoning is only performed on visual graph. }
            \begin{tabular}{lcccc}
            \hline
            Method & Backbone & GI & SC-loss & mIoU \\
            \hline
            baseline & Res50 &  &  & 48.5 \\
            +VisG    & Res50   &  &  &50.2 \\
            GINet & Res50 & \checkmark &  & 51.0 \\
            GINet & Res50 & \checkmark & \checkmark & 51.7 \\
            \hline
            baseline & Res101 &  &  & 51.4 \\
             +VisG    & Res101   &  &  & 53.0\\
            GINet & Res101 & \checkmark &  & 53.9 \\
            GINet & Res101 & \checkmark & \checkmark & 54.6 \\
            \hline
            \end{tabular}
         \label{ablation}
    \end{minipage}\quad
    \begin{minipage}{0.45\linewidth}
     \centering
          \makeatletter\def\@captype{table}\makeatother\caption{Comparisons of accuracy and efficiency with other methods. All experiments are based on ResNet50. ``Para'' represents the extra parameters relative to the backbone. ``FPS'' indicates the inference speed of models.}
         \begin{tabular}{lccc}
         \hline
         Methods & mIoU & Para (M) & FPS \\
         \hline
         baseline & 48.5 & 9.5 & 48.6 \\
         +GCU \cite{li2018beyond} & 50.4 & 11.9 & 40.3 \\
         +GloRe \cite{li2019graph} & 50.2 & 11.2 & 45.3 \\
         +PSP \cite{zhao2017pyramid} & 50.4 & 23.1 & 37.5 \\
         +ASPP \cite{chen2017rethinking} & 49.4 & 16.1 & 28.2 \\
         \hline
         GINet(ours)  & 51.7 & 10.2 & 46.0 \\
         \hline
         \end{tabular}
         \label{FPS}
   \end{minipage}
  \end{figure} 
\subsection{Experiments on Pascal-Context}
We first conduct experiments on the Pascal-Context dataset with different settings and compare our GINet with other popular context modeling methods.
Then we show and analyze the visualization results of our GINet.
Finally, we compare with state-of-the-art to validate the efficiency and effectiveness of our method.
   
\subsubsection{Ablation Study}

First, we show the effectiveness of the proposed GI unit and SC-loss. 
Then, we compare our method with other popular context modeling methods and study the influence of SC-loss in terms of weight.
Finally, different word embedding approaches are tested to show the robustness of our proposed method. 
  
{\bf Effectiveness of GI unit and SC-loss} \ We design a detailed ablation study to verify the effect of our GI unit and SC-loss.
Specifically, FCN with Joint Pyramid Upsampling Module \cite{wu2019fastfcn} is chosen as our baseline model.
As shown in Table 1, the baseline model achieves 48.5\% mIoU. By performing reasoning over the VisG (row2), there is an improvement in performance by 1.7\% (50.2 v.s. 48.5).
Instead, by adopting our GI unit upon the baseline model to capture context information from both visual regions and linguistic knowledge, one can see from Table \ref{ablation} (row 3), there is a significant increase in performance by 2.5\% (51.0 v.s. 48.5), which demonstrates the effectiveness of the GI unit and our context modeling method.
Furthermore, by constraining the global information of semantic concepts, SC-loss can further improve the model performance to 51.7\% mean IoU.
Deeper pre-trained backbone provides better feature representations. GINet configured with ResNet-101 can obtain 54.6\% mIoU, which outperforms the baseline mode by 3.2\% in terms of mIoU.


\begin{figure}[t]
   \centering
   \begin{minipage}[t]{0.48\linewidth}
   \centering
   \includegraphics[width=5cm]{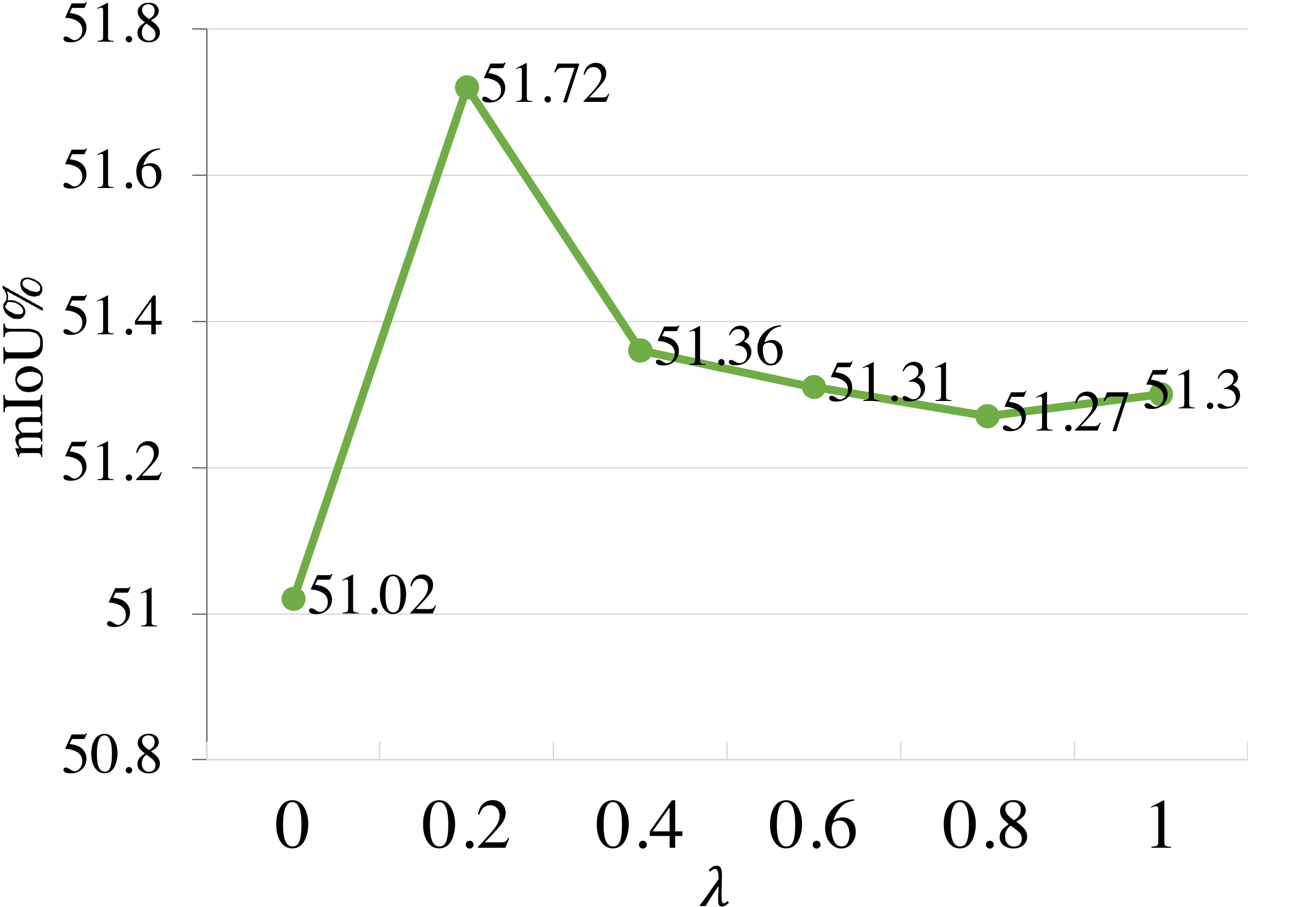}
   \caption{Ablation study on the weight of SC-loss. GINet can achieve the best results with $\lambda = 0.2$.}
   \label{fig:sc-loss}
   \end{minipage}\quad
   \begin{minipage}[t]{0.48\linewidth}
   \centering
   \includegraphics[width=5cm]{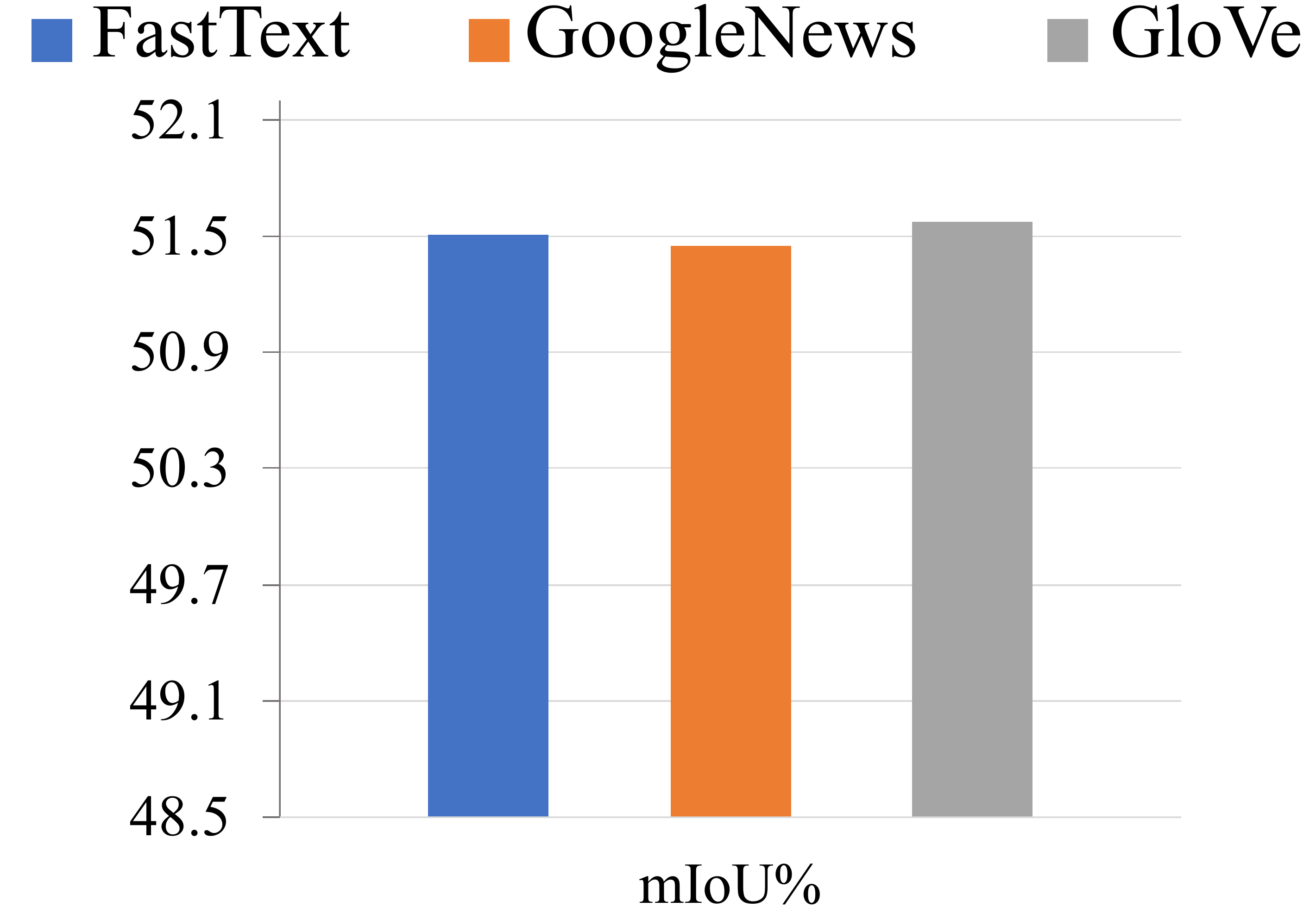}
   \caption{Ablation study on different word embedding methods. ({\em Best viewed in color.})}
   \label{fig:word}
   \end{minipage}
   \end{figure}
   

{\bf Comparisons with context modeling methods} 
Firstly, we compare the GINet with VisG-based context reasoning methods i.e., GCU \cite{li2018beyond}, GloRe \cite{chen2019graph}.
GloRe, GCU uses the typical projection, reasoning, and back-projection methods to model spatial context information.
To ensure the fairness, we reproduce these methods.
We report the model performances in terms of mIoU. As shown in Table \ref{FPS},
Compared to GCU's 50.4\% and GloRe's 50.2\%, our GINet achieves the highest score of 51.7\% mIoU. 
This proves the effectiveness of introducing linguistic knowledge and label dependency upon the visual image region reasoning.
PSPNet \cite{zhao2017pyramid} and DeepLab\cite{chen2017rethinking} are classic methods for constructing visual context information, and their performance is lower than our GINet.
To further analyze the efficiency of these context modeling methods, we list the inference speed (frame per second, denoted as FPS) of these models. 
FPS is measured on a Tesla-V100 GPU with input size 512 $\times$ 512. As shown in Table \ref{FPS}, our model achieves 46.0 FPS, which outperforms all other context modeling methods.


{\bf Importance of the Weights of the SC-loss} \ In order to study the necessity and effectiveness of the SC-loss, we train our GINet using different weights for the SC-loss, \emph{e.g.}, $\lambda$= \{0.2, 0.4, 0.6, 0.8, 1.0\}.
It is worth noting that $\lambda=0$ means that the SC-loss is not applied.
As shown in Figure~\ref{fig:sc-loss}, SC-loss can effectively improve the model performance when $\lambda=0.2$.
In our experiments, higher weights don't bring more performance increase.

{\bf GINet with different word embedding} \ 
By default, we use GloVe \cite{pennington2014glove} as the initial representation of SemG. 
To verify the robustness of our method and see the influence of using different word embedding representations, we conduct experiments by applying three popular word embedding methods, \emph{e.g.}, FastText \cite{joulin2016fasttext}, GoogleNews \cite{mikolov2013efficient} and GloVe \cite{pennington2014glove}.
As shown in Figure~\ref{fig:word}, there is no significant performance fluctuation using different word embedding representations. 
This observation suggests that our method is quite general. No mather what word mebedding methods are used, the proposed approach can capture the semantic information effectively.

\begin{figure}[t]
   \begin{center}
   \includegraphics[width=0.95\linewidth]{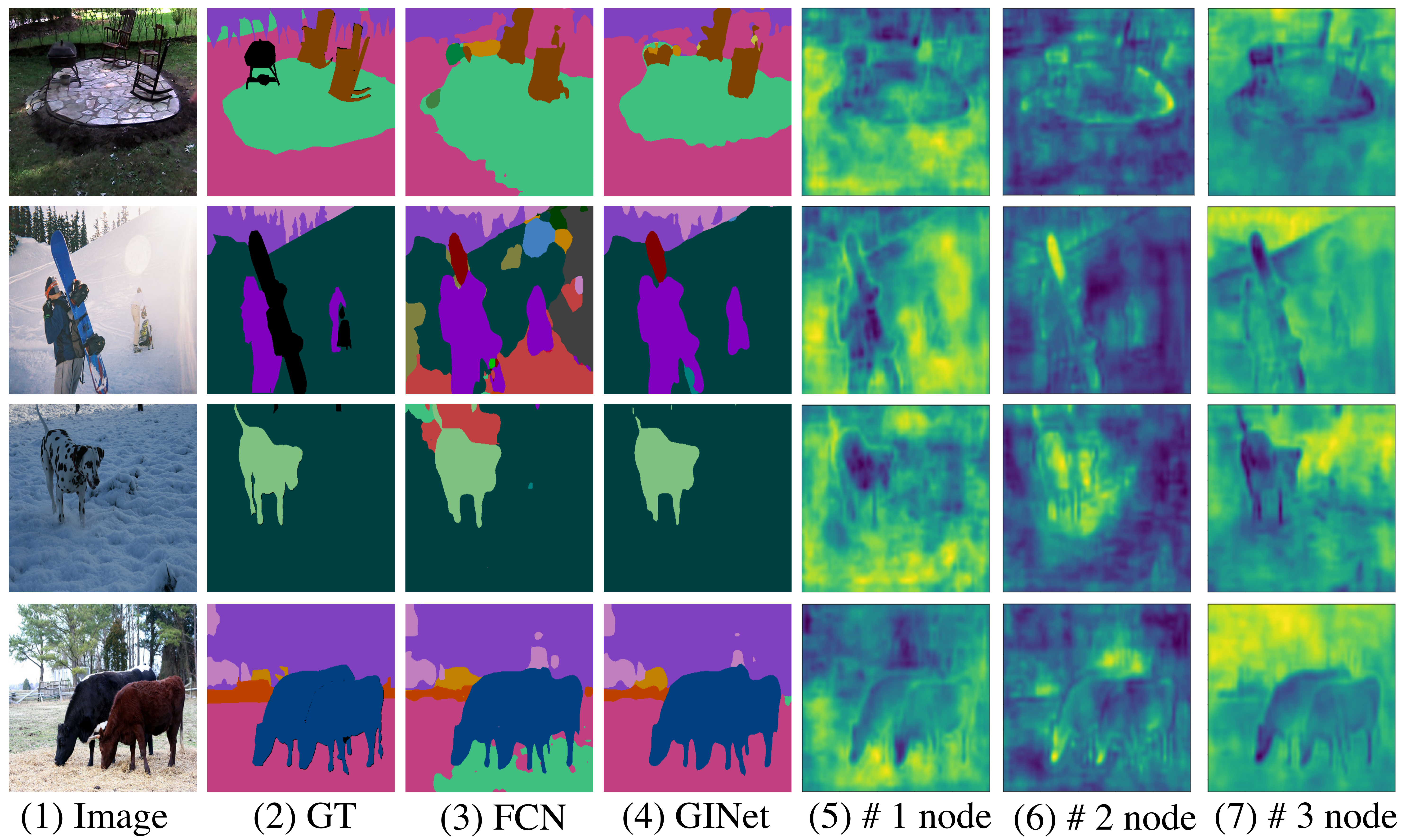}
   \end{center}
      \caption{Visualization of results and weights of projection matrices, all examples are from the Pascal-Context validation dataset.({\em Best viewed in color.})}
   \label{fig:visualization}
   \end{figure}

\subsubsection{ Visualization and Analysis}
   
In this section, we provide a visualization of scene parsing results and projection matrix. 
Then we analyze the qualitative results delivered by the proposed method.

The scene parsing results are shown in Figure \ref{fig:visualization}. 
Specifically, the first and second columns list the RGB input images and the ground truth scene parsing images, respectively. 
We compare baseline FCN \cite{Long_2015_CVPR} with our method in column 3 and column 4. 
One can see from the predicted parsing results images that our method shows considerable improvements. 
Particularly, it can be seen from rows 2 and 3, where the snow scene changes significantly in texture and color due to the illumination variation in the second and the fourth examples.
By incorporating the semantic graph to promote the reasoning over the visual graph, our method successfully obtained more accurate parsing results.
In the fourth row, it is fairly difficult to distinguish the green and yellow grass in the image only by the spatial context, while our method still identified the object correctly by incorporating semantic information, where the color changes mislead the FCN method.

Moreover, we can show that our method aggregates similar features from the visual feature map into a node in the visual graph. 
The graph nodes learn rich representations of different regions, and reasoning on these nodes can effectively capture relationships of image regions.
We select three nodes (marked as \#1, \#2 and, \#3) and show their corresponding projection weights in columns 5, 6, and 7, respectively.
It can be observed that different nodes correspond to relevant regions in the image (the brighter areas in the image means high response).
It can be seen from the 2nd row in Figure \ref{fig:visualization}, node \#1 aggregates and corresponds more with the background areas, while node \#2 highlights the main objects in the images and node \#3 shows more responses to the sky area in this example.


\setlength{\tabcolsep}{4pt}
\begin{table}[t]
\begin{center}
 \caption{Comparison with the state-of-the-art approaches on PASCAL-Context dataset, COCO stuff test set, and ADE20K validation set. ``${\dagger}$'' means the model has been pre-trained on COCO Stuff. ``\_" means no public results available. ``${*}$'' means employing online hard example mining(OHEM\cite{Shrivastava_2016_CVPR}).}
\label{pascal}
\begin{tabular}{lcccc}
\hline\noalign{\smallskip}
\multirow{2}*{\textbf{Method}}                        & \multirow{2}*{\textbf{Backbone}}                           & \multicolumn{3}{c}{\textbf{mIoU}\%} \\
\cline{3-5}\noalign{\smallskip}
~                                                                     &  ~                                                      &     PASCAL-Context    &COCO Stuff   & ADE20K\\
\noalign{\smallskip}
\hline
\noalign{\smallskip}
   CCL \cite{ding2018context}                         & ResNet-101                                        &  51.6                                 & 35.7             &    -                 \\
   PSPNet \cite{zhao2017pyramid}                 & ResNet-101                                        & 47.8                              &  -                   &43.29 \\
   EncNet \cite{zhang2018context}                 & ResNet-101                                        & 51.7                               &  -                  &44.65  \\
   TKCN \cite{wu2019tree}                              & ResNet-101                                       & 51.7                              &-                     &-     \\
   CFNet \cite{wu2019consensus}                  & ResNet-101                                       & 52.4                              &36.6                    &-     \\
   DUpsampling \cite{tian2019decoders}        & Xception-71                                        & 52.5                             &-                      & - \\
   SGR$\dagger$ \cite{liang2018symbolic}    & ResNet-101                                         & 52.5                             &39.1                &44.32 \\
   DSSPN \cite{liang2018dynamic}                & ResNet-101                                        &    -                                  & 37.3                &43.68  \\
   DANet \cite{fu2019dual}                              & ResNet-101                                        & 52.6                             &39.7                 &-\\
   ANN$^{*}$ \cite{zhu2019asymmetric}         & ResNet-101                                       & 52.8                               &  -                      &45.24  \\
   FastFCN \cite{wu2019fastfcn}                    & ResNet-101                                       & 53.1                               & -                       &44.34\\
   GCU \cite{li2018beyond}                           & ResNet-101                                        &  -                                    &  -                       &44.81 \\
   EMANet \cite{li2019expectation}               & ResNet-101                                       & 53.1                               &39.9                 &-\\
   SVCNet \cite{ding2019semantic}             &ResNet-101                                         & 53.2                               &39.6                  &-\\
   CCNet${*}$ \cite{huang2018ccnet}        & ResNet-101                                      &      -                                 &    -                        & 45.22 \\
   DMNet \cite{He_2019_ICCV}                    & ResNet-101                                    &   54.4                                     &   -                         &45.50 \\
   ACNet${*}$ \cite{fu2019adaptive}                  &ResNet-101                                          &  54.1                              &40.1                    &\textbf{45.90} \\
   \hline
   GINet (Ours)                                          & ResNet-101                                        &  \textbf{54.9}                    & \textbf{40.6}        &45.54     \\
 \hline
\end{tabular}
\end{center}
\end{table}
\setlength{\tabcolsep}{1.4pt}


%
%

\subsubsection{Comparisons with state-of-the-art methods}

We report 60 categories performance (including background) to compare with the state-of-the-art methods.
As shown in Table~\ref{pascal}, our GINet achieves the best performance and outperforms the SOTA DMNet\cite{He_2019_ICCV} by 0.5\%, which shows that our method is truly competitive.
DMNet incorporates multiple Dynamic Convolutional Modules to adaptively exploit multi-scale filters to handle the scale variation of objects.
In addition, the ACNet \cite{fu2019adaptive} obtains 54.1\% mIoU, which captured the pixel-aware contexts by a competitive fusion of global context and local context according to different per-pixel demands.
However, our method extracts the semantic representation from the visual features under the guidance of the general semantic graph,
and construct a semantic centroid to get the similarity score for each category.


\subsection{Experiments on COCO Stuff}

To further demonstrate the generalization of our GINet, we also conduct experiments on the COCO Stuff dataset \cite{caesar2018coco}.
Comparisons with state-of-the-art methods are shown in Table~\ref{pascal}.
Remarkably, the proposed model achieves 40.6\% in terms of mIoU, which outperforms the best methods by a large margin.
Among the current state-of-the-art methods, 
ACNet \cite{fu2019adaptive} introduced a data-driven gating mechanism to capture global context and local context according to pixel-aware context demand,
DANet \cite{fu2019dual} deployed the self-attention module to capture long-range contextual information, and
EMANet \cite{li2019expectation} proposesd the EMA Unit to formulate the attention mechanism into an expectation-maximization manner.
In contrast to these methods, our GINet considers the operation of capturing long-range dependencies as the way of graph reasoning, 
and additionally introduces the semantic context to enhance the discriminant property for the features.

\subsection{Experiments on ADE20K}

Finally, we compare our method and conduct experiments on the ADE20K dataset \cite{Zhou_2017_CVPR}.
Table \ref{pascal} compares the GINet performance against state-of-the-art methods. Our GINet outperforms the prior works and sets the new state-of-the-art mIoU to 45.54\%.
It is noting that our result is obtained by a regular training strategy in contrast to ANN \cite{zhu2019asymmetric}, CCNet \cite{huang2018ccnet} and ACNet \cite{fu2019adaptive} where OHEM\cite{Shrivastava_2016_CVPR} is applied to help cope with difficult training cases.
The ANN proposes an asymmetric fusion of non-local blocks to explore the long-range spatial relevance among features of different levels.
The CCNet used a recurrent criss-cross attention module that aggregates contextual information from all pixels.
We emphasize that achieving such an improvement on the ADE20K dataset is hard due to the complexity of this dataset.

\section{Conclusion}

We have presented a graph interaction unit to promote contextual reasoning over the visual graph by incorporating the semantic knowledge.
We have also developed a Semantic Context loss upon the semantic graph output of graph interaction unit to emphasize the categories that appear in the scene and suppress those do not appear in the scene.
Based on the proposed graph interaction unit and Semantic Context loss, we have developed a novel framework called Graph Interaction Network (GINet).
The proposed approach based on the new framework outperforms state-of-the-art methods by a significant gain in performance on two challenging scene parsing benchmarks, e.g., Pascal-Context and COCO Stuff, and achieves a competitive performance on ADE20K dataset.

\clearpage


%
%
\bibliographystyle{splncs04}
\bibliography{egbib}
\end{document}